\documentclass[a4paper, 10pt, conference]{IEEEtran}
\IEEEoverridecommandlockouts

\usepackage{stfloats}
\usepackage{makecell}
\usepackage{url}
\usepackage{cite}
\usepackage{amsmath,amssymb,amsfonts}
\usepackage{algorithmic}
\usepackage{graphicx}
\usepackage{textcomp}
\usepackage{xcolor}
\def\BibTeX{{\rm B\kern-.05em{\sc i\kern-.025em b}\kern-.08em
    T\kern-.1667em\lower.7ex\hbox{E}\kern-.125emX}}
\begin{document}

\title{TiBERT: Tibetan Pre-trained Language Model*\\
\thanks{Supported by National Nature Science Foundation (No. 61972436).}
}

\author{
    \IEEEauthorblockN{Yuan Sun$^{1,2*}$, Sisi Liu$^{1,2}$, Junjie Deng$^{1,2}$, Xiaobing Zhao$^{1,2}$}
    \IEEEauthorblockA{$^1$ Minzu University of China, China}
    \IEEEauthorblockA{$^2$ National Language Resource Monitoring \& Research Center Minority Languages Branch}
    \IEEEauthorblockA{tracy.yuan.sun@gmail.com, liusisi.s@qq.com, 1106130700@qq.com, nmzxb\_cn@163.com}
    }

\maketitle

\begin{abstract}
The pre-trained language model is trained on large-scale unlabeled text and can achieve state-of-the-art results in many different downstream tasks. However, the current pre-trained language model is mainly concentrated in the Chinese and English fields. For low resource language such as Tibetan, there is lack of a monolingual pre-trained model. To promote the development of Tibetan natural language processing tasks, this paper collects the large-scale training data from Tibetan websites and constructs a vocabulary that can cover 99.95$\%$ of the words in the corpus by using Sentencepiece. Then, we train the Tibetan monolingual pre-trained language model named TiBERT on the data and vocabulary. Finally, we apply TiBERT to the downstream tasks of text classification and question generation, and compare it with classic models and multilingual pre-trained models, the experimental results show that TiBERT can achieve the best performance. Our model is published in \url{http://tibert.cmli-nlp.com/}.
\end{abstract}

\begin{IEEEkeywords}
Pre-trained language model, Tibetan, Sentencepiece, TiBERT, Text classification, Question generation
\end{IEEEkeywords}

\section{Introduction}
Pre-trained language models represented by BERT have changed the research paradigm of natural language processing. These models can be pre-trained on large-scale unlabeled corpora to obtain rich contextual representations, and through transfer learning, downstream tasks with small-scale labeled datasets can also achieve better performance, which solves the problem of less labeled data in low-resource languages. Therefore, pre-trained langage models are crucial for low-resource languages such as Tibetan.

At present, monolingual pre-trained language models are mainly concentrated in the Chinese and English fields, and low resources have not been fully applied. To solve this problem, Google released a multilingual model \cite{b1}. Multilingual pretrained model can process multiple languages simultaneously and provide support for downstream tasks in multiple low-resource languages. However, this paper analyzes the current multilingual pre-training models, including mBERT  \cite{b1}, XLM-RoBERTa \cite{b2} and T5 \cite{b3} models, etc. The multilingual models select the top 104 languages used in Wikipedia for model training, and we also find that the training data of these multilingual pre-training models did not include the minority languages such as tibetan, which seriously hindered the development of various downstream tasks in Tibetan. 

For a long time, due to the difficulty of acquiring minority language corpus in China, there is no public dataset, and the relative research is relatively scarce. Recently, the Joint Laboratory of HIT and iFLYTEK Research (HFL) released the first multi-language pre-trained model named CINO (Chinese mINOrity pre-trained language model) for minority languages , this model can promote the research of natural language processing tasks in ethnic minority languages and dialects such as Tibetan, Mongolian, Uyghur, Kazakh, Korean, Zhuang, Cantonese and so on. The emergence of CINO greatly promotes the research and development of Chinese minority language information processing. However, the effect on downstream tasks is still far from the Chinese and English field.

To further promote the development of various downstream tasks of Tibetan natural language processing, and obtain the results obtained by BERT in English and Chinese, this paper trains a monolingual language model named TiBERT for Tibetan. We crawl the Tibetan corpus from the websites of Tibet People’s Network and Qinghai Provincial People’s Government Network as the training data of TiBERT, and use the Sentencepiece \cite{b4} model to segment the Tibetan sub-words. To verify the performance of TiBERT, this paper evaluates TiBERT in two downstream tasks of text classification and question generation. The main contributions of this paper are as follows:

(1)To better express the semantic information of Tibetan and reduce the problem of OOV, this paper uses the unigram language model of Sentencepiece to segment Tibetan words and constructs a vocabulary that can cover 99.95$\%$ of the words in the corpus.

(2)To further promote the development of various downstream tasks of Tibetan natural language processing, this paper collected a large-scale Tibetan dataset and trained the monolingual Tibetan pre-trained language model named TiBERT.

(3)To evaluate the performance of TiBERT, this paper conducts comparative experiments on the two downstream tasks of text classification and question generation. The experimental results show that the TiBERT is effective.

\section{Related Work}
Word embedding can convert words in natural language into dense vectors recognized by computers, and similar words will have similar vector representations. Word embedding can dig out the features between words and sentences in the text, which is important for natural language processing tasks. Previous word representation approaches, such as fastText \cite{b5}, word2vec \cite{b6}, and GloVe \cite{b7}, can just learn a simple and specific vector embedding from the data. The word embedding of each word is static and non-contextualized embedding, we cannot distinguish multiple meanings of a word. To solve this problem, some pre-trained language models are trained to generate contextual word vector representations. ELMo \cite{b8} is the first to propose and successfully apply a contextualized word embedding, which uses a bidirection LSTM network to obtain word embeddings based on context. The essence of ELMo is to first use the language model to learn the word representations of words on a large corpus, and the word representations are non-contextualized, then use the training data to fine-tune the pre-trained ELMo model, we can obtain the word vector representation of the word in the current context by using the context information of the training data. Another notable model is ULMFit \cite{b9}, which is based on a LSTM architecture and a language modeling task. The ULMFit includes three stages: general-domain LM pretraining, Target task LM ﬁne-tuning and Target task classiﬁer ﬁne-tuning. Since then, the structure of pre-training and fine-tuning has attracted more attention of natural language processing community.

\begin{table*}[bp]
\caption{The sentences after sub-word lever segmentation and masked}
\begin{center}
\begin{tabular}{cc}
 \hline
     \makecell[c]{Sentence}&\makecell[c]{\\\includegraphics[width=0.4\textwidth]{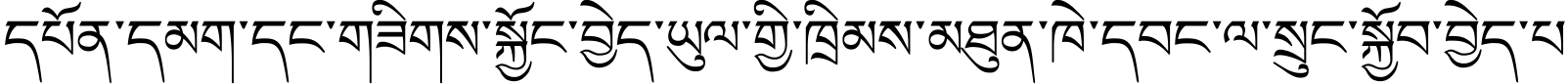}\\
     (Protect the legitimate rights and interests of officers and soldiers and receiving \\preferential treatment.)}\\
      \hline
      \makecell[c]{Segmentation} & \makecell[c]{\\\includegraphics[width=0.4\textwidth]{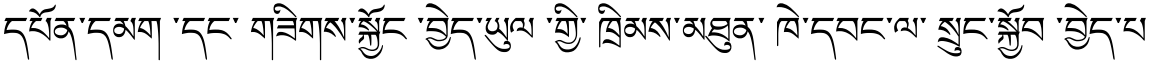}\\}\\
      \hline
      \\
      \makecell[c]{Masked} & \makecell[c]{[CLS][MASK]\includegraphics[width=0.4\textwidth]{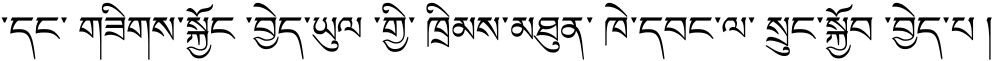}[SEP]\\
([CLS] Protect the legitimate rights and interests of [MASK] and receiving \\preferential treatment [SEP])}
\\
\hline
\end{tabular}
\label{tab1}
\end{center}
\end{table*}

The performance of the model on many tasks has been greatly improved by processing the polysemy of words. In particular, after the introduction of the unsupervised pre-training methods \cite{b10,b8} in Transformer architecture\cite{b11}, various pre-training models began to emerged, such BERT\cite{b10}, ALBERT\cite{b12}, RoBERTa\cite{b13}, GPT \cite{b14}, GPT2 \cite{b15}, T5 \cite{b3} and XLNet\cite{b16}. Among them, BERT is the most popular of various pre-trained language models and is particularly influential, which can jointly adjust the left and right contexts of all layers to pre-train deep bidirectional representations in unlabeled text and alleviates the unidirectionality constraint. it has established state-of-the-art results for English in most downstream tasks of natural language processing. However, the success of pre-trained monolingual language model and its variants is largely limited to English and Chinese. 

For the low-resource languages, available model is the multilingual model \cite{b2,b1,b3,b17}, Although the multilingual models have remarkable cross-language generalization capabilities, several studies have also shown that the available monolingual BERT model is significantly better than the multilingual BERT model. Thus, a few monolingual BERT-based models have been released, such as for Spanish \cite{b18}, French \cite{b19,b20}, Italian \cite{b21} German, Finnish \cite{b22}, and so on. They all achieved good performance in downstream tasks. However, in our knowledge, only CINO\cite{b17} can be used for the downstream task research of Tibetan. In order to further explore the pre-trained language model of Tibetan, this paper constructs a monolingual Tibetan pre-trained language model named TiBERT and compare with CINO.

\section{TiBERT Model}

\subsection{Pretraining Data}

\subsubsection{Data Collection}
There is currently no publicly available large-scale corpus of Tibetan, this paper obtains original Tibetan data from 21 Tibetan websites including Tibet People’s Network and Qinghai Provincial People’s Government Network. The data contains knowledge in various fields such as current affairs, economy, technology, society, law, sports, life, nature, culture, geography, art, military, education, history, and people. We clean and filter the original data, discarded the non-text information in the article, such as pictures, links, special symbols, etc., and selected articles with more than 100 words. Finally, we collect 3.56G of training data. After that, all the text needs to be segmented into sentences \cite{b23}.

\subsubsection{Vocabulary Construction}
To construct Tibetan vocabulary and solve the problem of OOV (out-of-vocabulary words), this paper segment the Tibetan data at the word level, syllable level, and sub-word level respectively, and statistical the size of the vocabulary generated after the segmentation.

(1) Word level segmentation

When we segment the data at the word level, the size of the vocabulary exceeds 100,000. The vocabulary is too large, and increases the amount of computation of the machine and takes longer time and more computing resources to train the model.

(2) Syllable level segmentation

Tibetan is a pinyin language. The smallest unit of Tibetan is a syllable, which contains one or up to seven characters. The syllables are separated by ".", so we use "." to segment Tibetan data, and for words which cannot be segmented, such as the year and time in Tibetan, they are segmented according to a single symbol. We select the high-frequency words in the data to construct the vocabulary. When the frequency threshold is set to 25, the size of vocabulary has reached 35,162, and according to our statistics, the number of [UNK] accounts for 15$\%$ of all the dataset. If we use syllable level segmentation, we also need to construct a very large vocabulary to reduce OOV, which will increase the training time.

(3) Sub-word level segmentation

The word level and syllable level segmentation will cause the generated vocabulary to be too large, which will affect the training efficiency. Therefore, we need to extract a more coarse-grained vocabulary. Sentencepiece treats the original Tibetan sentences as a sequence of “Unicode” characters, and use statistical learning algorithm to generate word segmentation model and vocabulary of specified size, which is very suitable for the study of Tibetan. Based on this, we can train a word segmentation model that meets the requirements on the unlabeled dataset. Sentencepiece provides four modes: bpe, unigram, char, and word. The bpe model can only generate a unique sub-word sequence for a sentence, while the unigram language model can generate multiple candidate sub-word sequences, which can make the model more robust to noise and sub-word segmentation errors. Therefore, this paper uses the unigram language model to generate a vocabulary which can cover 99.95$\%$ of the characters in the dataset. Finally, the Tibetan vocabulary we constructed contains 30,005 words. We use the model of Sentencepiece to segment Tibetan sentences, and results are shown in Tab ~\ref {tab1}.

\begin{table}[htbp]
\caption{TiBERT parameters}
\begin{center}
\setlength{\tabcolsep}{10mm}{
\begin{tabular}{cc}
\hline
\makecell[c]{Parameters} & \makecell[c]{Values}\\
\hline
\makecell[c]{hidden$\_$dropout$\_$prob} & \makecell[c]{0.1}\\
      \makecell[c]{hidden$\_$size} & \makecell[c]{768}\\
      \makecell[c]{intermediate$\_$size} & \makecell[c]{3,072}\\
      \makecell[c]{max$\_$position$\_$embeddings} & \makecell[c]{512}\\
      \makecell[c]{num$\_$attention$\_$heads} & \makecell[c]{12}\\
      \makecell[c]{num$\_$hidden$\_$layers} & \makecell[c]{12}\\
      \makecell[c]{vocab$\_$size} & \makecell[c]{30,005}\\
      \hline
\end{tabular}}
\label{tab2}
\end{center}
\end{table}

\begin{table*}[bp]
\caption{Performances on title classification}
\begin{center}
\setlength{\tabcolsep}{5mm}{
\begin{tabular}{ccccc}
\hline
     Model & Accuracy($\%$) & macro-Precision($\%$) & Macro-Recall($\%$) & macro-F1($\%$)\\
      \hline
      CNN(letter) & 47.97 & 39.57 & 38.63 & 38.03\\
      CNN(syllable) & 54.42 & 49.22 & 48.34 & 48.64\\
      Transformer & 46.88  &54.21 & 46.88 & 42.91\\
      TextCNN  &60.94  &59.58 & 60.94 & 58.90\\
      DPCNN   &64.06 & 59.05 & 64.06 & 59.68\\
      TextRCNN  &65.62 & 63.12 & 65.62 & 60.80\\
      TiBERT & \textbf{65.62} & \textbf{62.88} & \textbf{65.62} & \textbf{61.72}\\
      TiBERT+CNN & 65.62 & 59.47 & 65.62 & 60.93\\
      \hline
\end{tabular}}
\label{tab3}
\end{center}
\end{table*}

\subsection{Models}
We use the same architecture of BERT to train TiBERT, including a multi-layer bidirectional Transformer, the model size is 12 layers, 768 hidden dimensions, 12 attention heads and 110M parameters. The original BERT contains two supervision tasks: (1) MLM (masked language model), to train a deep bidirection representation, for a given sequence, we randomly select 15$\%$ of tokens to replace, of which 80$\%$ replaced by <mask>, 10$\%$ is randomly replaced with other tokens, and 10$\%$ remains unchanged. (2) NSP (next sentence prediction), for a given pair of input sentences A and B, the model learns to predict whether sentence B is the next sentence after A. We employed the MLM and NSP objective in TiBERT.

This paper uses Sentencepiece to mark the corpus, the input token of the model contains words and sub-words. Studies have shown that masking whole words instead of individual sub-words can improve the performance of the model \cite{b24}. The sentences are masked by masked language model is shown in Tab \ref{tab1}, and the parameters of the TiBERT model are shown in Tab~\ref{tab2}.

\section{Evaluation of TiBERT} 

We use the two downstream tasks of Tibetan text classification and question generation to verify the performance of TiBERT.

\subsection{Text Classification}

This paper uses the Tibetan News Classification Corpus (TNCC) \cite{b25} released by the Natural Language Processing Laboratory of Fudan University for text classification. The dataset collected from China Tibet online website, which contains 12 distinct categories including politics, Economics, education, tourism, environment, language, literature, religion, arts, medicine, customs, and Instruments. The dataset has a total of 9,203 news. We divide the dataset into a training set, a development set and a test set. The training set accounts for 80$\%$, and both the development set and the test set account for 10$\%$.

To verify the effects of the TiBERT model on short text and long text classification respectively, we conduct text classification experiments on the title and document respectively. The calculations of evaluation methods are shown in Equation (1)-(4).
\begin{equation}
    \text { Accuracy } = \frac{T P+T N}{T P+T N+F P+F N}
\end{equation}
\begin{equation}
    \text { Precision } = \frac{T P}{T P+F P}
\end{equation}
\begin{equation}
    \text { Recall }=\frac{T P}{T P+F N}
\end{equation}
\begin{equation}
    F 1=\frac{2 * \text { Precision } * \text { Recall }}{\text { Precision }+\text { Recall }}
\end{equation}
where TP is the number of the true category and predicted category are positive examples, FP is the number of cases where the true category is negative but the predicted category is positive, FN is the number of cases where the true category is positive but the predicted category is negative, TN is the number of the true category and predicted category are negative examples. We use macro-averaging to evaluate multi-classification tasks, that is, to calculate the accuracy, recall, and F1 of each category, and then calculate the average value to get macro-precision, macro-recall and macro-F1.

\begin{table*}[htbp]
\caption{Performances on document classification}
\begin{center}
\setlength{\tabcolsep}{5mm}{
\begin{tabular}{ccccc}
 \hline
     Model & Accuracy($\%$) & Macro-Precision($\%$) & Macro-Recall($\%$) & Macro-F1($\%$)\\
      \hline
      CNN(syllable) & 61.51 & 59.39 & 56.65 & 57.34\\
      CINO-large & - & - & - & 68.6\\
      Transformer& 28.63 & 41.21 & 28.63 & 28.79 \\
      TextCNN  &61.71 & 61.65 & 61.71 & 61.53\\
      DPCNN   &62.91 & 63.61 & 62.91 & 61.17\\
      TextRCNN  &63.67 & 64.37 & 63.67 & 62.81\\
      TiBERT & \textbf{71.04} & \textbf{71.20} & \textbf{71.04} & \textbf{70.94}\\
      TiBERT+CNN & 70.39 & 70.54 & 70.39 & 70.23\\
      \hline
\end{tabular}}
\label{tab4}
\end{center}
\end{table*}

\subsubsection{News Title Classification}
\begin{figure}[htbp]
\centerline{\includegraphics[scale=0.28]{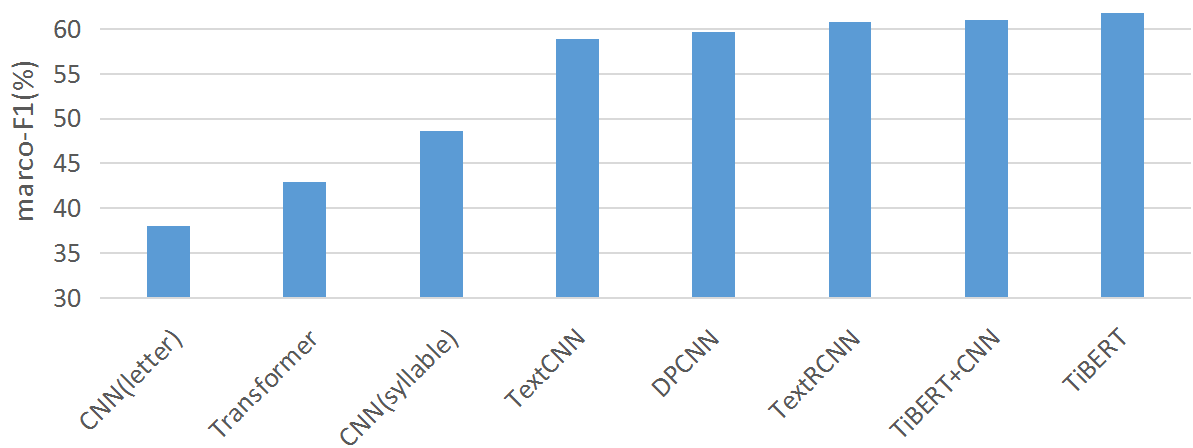}}
\caption{Performances on title classification.}
\label{fig1}
\end{figure}

We conduct an experiment on news title classification and use convolutional neural network (CNN), TextCNN \cite{b26}, DPCNN \cite{b27} TextRCNN \cite{b28}, Transformer \cite{b11} and TiBERT+CNN as the comparative experiment. The syllable is the basic unit of Tibetan, and the syllable in Tibetan contains very rich semantic information. Following the work of Qun et al.\cite{b25}, we can know that the classification effect of CNN at the Tibetan syllable level is better than the classification effect at the word level. Therefore, we choose the classification result of CNN at the Tibetan syllable level as the baseline model. The experimental results are shown in Tab~\ref{tab3} and Fig~\ref{fig1}. We can see that TiBERT has the best classification performance on the title classification, and reaching 61.72$\%$. On the TiBERT+CNN model, the performance will decline. It is speculated that CNN will destroy the original timing information of TiBERT and reduce its effect.

\begin{figure}[htbp]
\centerline{\includegraphics[scale=0.28]{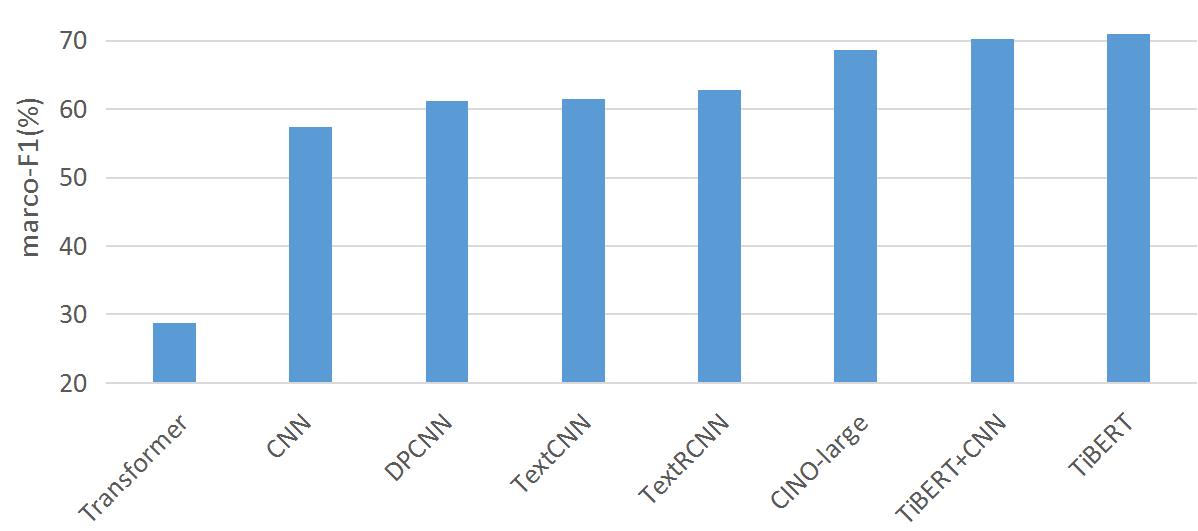}}
\caption{Performances on document classification.}
\label{fig2}
\end{figure}

\subsubsection{News Document Classification}
To verify the classification effect of TiBERT on long texts, this paper classifies documents and uses syllable-level CNN as the baseline model. Recently, the Joint Laboratory of HIT and iFLYTEK Research (HFL) released the first multi-language pre-trained model CINO (Chinese mINOrity pre-trained language model) for minority languages, and they announced the classification effect of CINO on the TNCC dataset, this paper compares the experimental result with the classification result of TiBERT, as shown in Tab~\ref{tab4} and Fig~\ref{fig2}. We can see that TiBERT also achieves the best performance in long text classification and reaches 70.94$\%$. The performance of CINO is worse than that of TiBERT, which matches the research results in other languages.

From Tab~\ref{tab3} and Tab~\ref{tab4}, we can see that the classification result of TiBERT on long text is better than that on short text, the main reason is that the long text contains more information, and the TiBERT model can learn more knowledge.

\subsection{Question Generation}
Question generation (QG) is the task of natural language generation, which takes a text and a target answer as input and automatically generate questions from the answer. The existing models mainly focus on recurrent neural networks (RNN) with attention mechanism and copy mechanism. Recently, researchers have begun to use pre-trained language models to guide question generation. This paper follows the work of Sun~\cite{b29} and uses the sequence-to-sequence model with attention and copy mechanism as the baseline model for question generation, The encoder uses a BiLSTM and a self-attention mechanism to encode paragraphs and answers to obtain contextual representations. The decoder uses a LSTM and copy mechanism to decode the output of the encoder.

\begin{table}[htbp]
\caption{The parameters of TiBERT-base question generation model}
\begin{center}
\setlength{\tabcolsep}{10mm}{
\begin{tabular}{cc}
\hline
      Parameters & Values\\
      \hline
      hidden$\_$size & 300\\
      embedding$\_$size & 768\\
      batch$\_$size & 64\\
      dropout &0.3\\
      learning rate & 0.1\\
      num$\_$epochs & 20\\
      vocab$\_$size & 30,005\\
      \hline
\end{tabular}}
\label{tab5}
\end{center}
\end{table}

\begin{table*}[bp]
\caption{Performances on question generation}
\begin{center}
\setlength{\tabcolsep}{5mm}{
\begin{tabular}{cccccc}
\hline
      Model & BLEU-1 & BLEU-2 & BLEU-3 & BLEU-4 & ROUGE-L\\
      \hline
      S2S$+$ATT$+$CP & 29.99 & 20.14 & 13.90 & 9.59 & 31.45\\
      \hline
      TiBERT & \textbf{35.48} & \textbf{28.60} & \textbf{24.51} & \textbf{21.30} & \textbf{40.04}\\
      \hline
\end{tabular}}
\label{tab6}
\end{center}
\end{table*}

\begin{table*}[bp]
\caption{Case analysis}
\begin{center}
\begin{tabular}{cccc}
\hline
      & Gold questions & S2S+ATT+CP & TiBERT\\
      \hline
      \makecell[c]{1} & \makecell[c]{\\\includegraphics[width=0.3\textwidth]{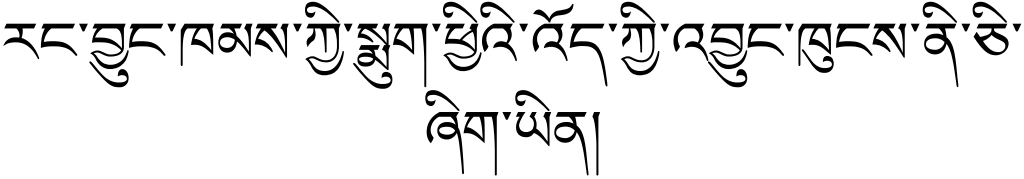}\\What is the source of \\ultraviolet light in nature?)} & \makecell[c]{\\\includegraphics[width=0.3\textwidth]{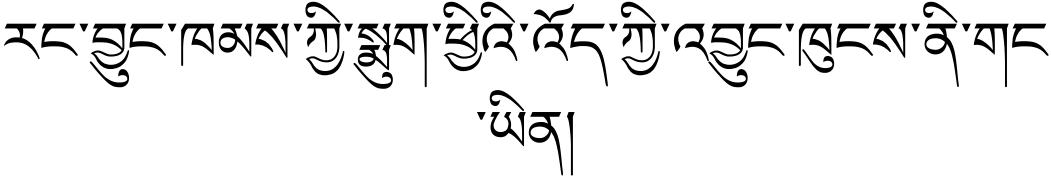}\\(What is the source of \\ultraviolet light in nature?)} & \makecell[c]{\\\includegraphics[width=0.3\textwidth]{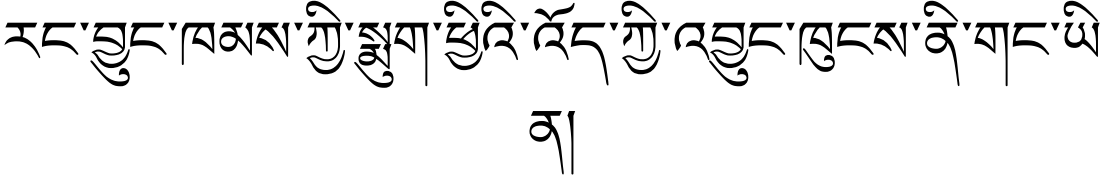}\\(What is the source of \\ultraviolet light in nature?)}\\
      
      \makecell[c]{2} & \makecell[c]{\\\includegraphics[width=0.3\textwidth]{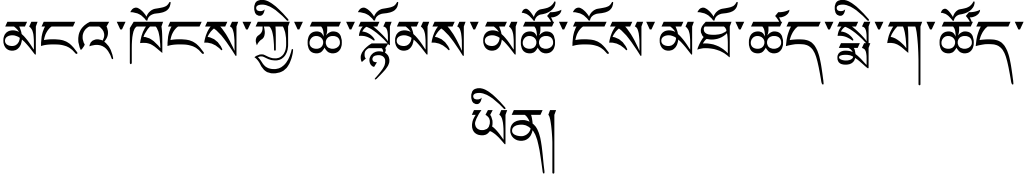}\\(What is the average \\elevation of the territory in meters?)} & \makecell[c]{\\\includegraphics[width=0.3\textwidth]{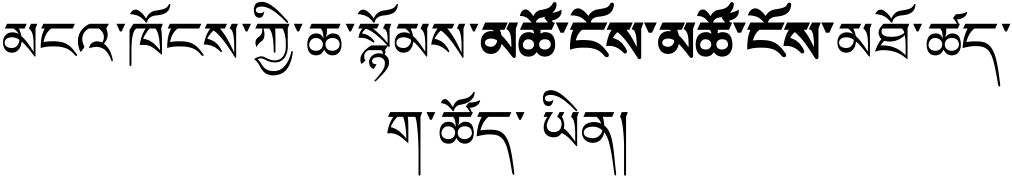}\\(What is the average\\ \textbf{elevation elevation} of the \\territory?)} & \makecell[c]{\\\includegraphics[width=0.3\textwidth]{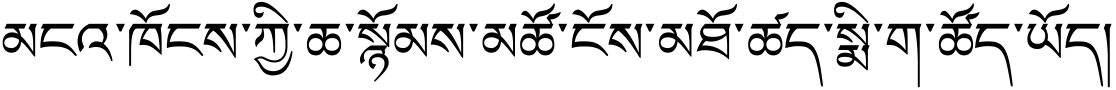}\\(What is the average elevation \\of the territory in meters?)}\\
   
      \makecell[c]{3} & \makecell[c]{\\\includegraphics[width=0.3\textwidth]{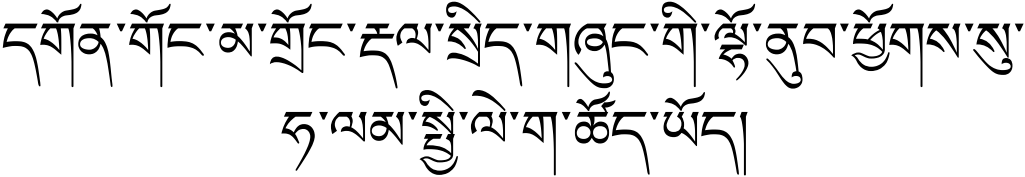}\\(How many kilometers southwest\\ of Gongma town government?)} & \makecell[c]{\\\includegraphics[width=0.3\textwidth]{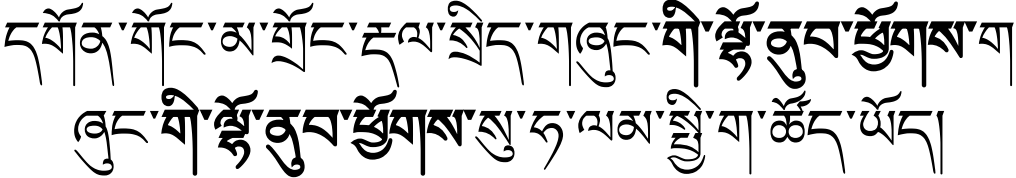}\\(How many centimeters is \\the \textbf{southwest southwest} of \\Gongma town government?)} & \makecell[c]{\\\includegraphics[width=0.3\textwidth]{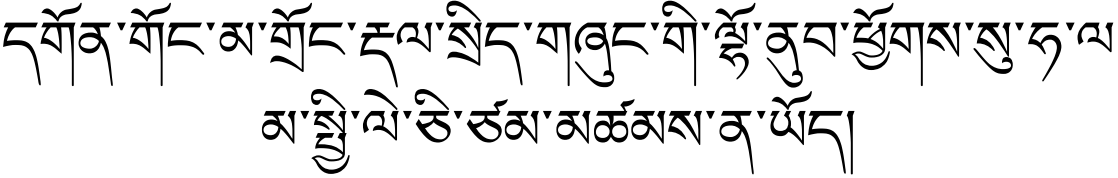}\\(How many kilometers \\southwest of Gongma town \\government?)}\\
  
      \makecell[c]{4} & \makecell[c]{\\\includegraphics[width=0.25\textwidth]{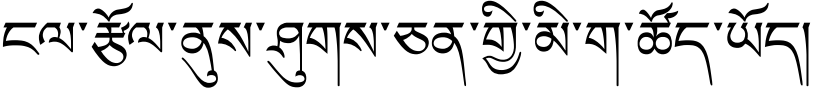}\\(How many people have the \\ability of work?)} & \makecell[c]{\\\includegraphics[width=0.2\textwidth]{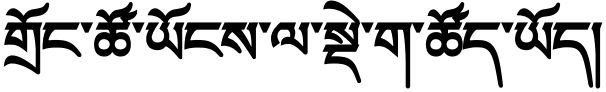}\\(\textbf{How many villages are}\\ \textbf{there in the whole town?})} & \makecell[c]{\\\includegraphics[width=0.3\textwidth]{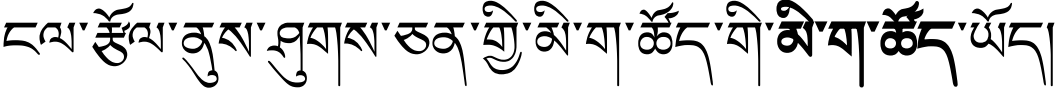}\\(How \textbf{many many} people\\ have the ability of work?)}\\
 
      \makecell[c]{5} & \makecell[c]{\\\includegraphics[width=0.25\textwidth]{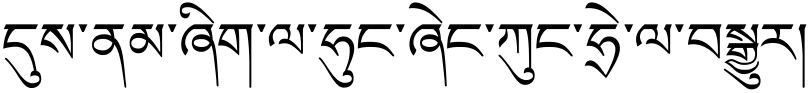}\\(When will it be changed to \\Red Star Commune?)} & \makecell[c]{\\\includegraphics[width=0.25\textwidth]{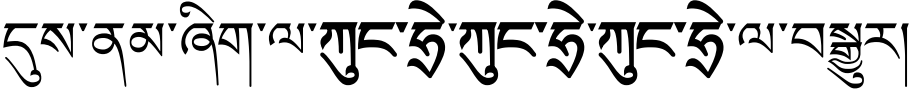}\\(When will it be changed to a\\ \textbf{commune commune} \\\textbf{commune}?)} & \makecell[c]{\\\includegraphics[width=0.25\textwidth]{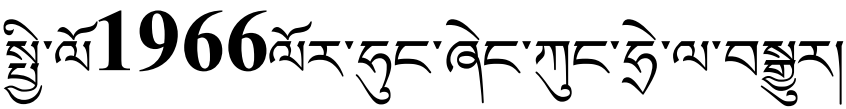}\\(Changed to Red Star \\Commune in \textbf{1966})}\\
      \hline
\end{tabular}
\label{tab7}
\end{center}
\end{table*}
\subsubsection{Dataset}
This paper uses TibetanQA dataset \cite{b30}, which is a Tibetan dataset for machine reading comprehension and contains 1,513 articles and 20,000 question and answer pairs. This dataset is the first high-quality Tibetan dataset for machine reading comprehension. We divide the data set into training set, development set and test set according to the ratio of 8:1:1.

To better evaluate the performance of the model, we use BLEU and ROUGE-L\cite{b31} as the metrics. The calculation is shown in Equation (5)-(9).

\begin{equation}
    B L E U=B P * \exp \left(\sum_{n=1}^{N} W_{n} \log P_{n}\right)
\end{equation}
\begin{equation}
    B P=\left\{\begin{array}{ll}
1 & l_{c}<l_{s} \\
e^{1-\frac{l_{s}}{l_{c}}} & l_{c} \geq l_{s}
\end{array}\right.
\end{equation}

where BP is the penalty factor, $W_{n}$ is the weight of the n-gram, $P_{n}$ is the accuracy of the n-gram, $l_{c}$ is the length of the question generated by the model, and $l_{s}$ is the length of the gold question. BLEU-2 uses the 2-gram language model to match gold answers.

The L in ROUGE-L refers to the longest common subsequence (LCS). The calculation of ROUGE-L uses the longest common subsequence of the generated question Q and the reference question Y.

\begin{equation}
    R_{l c s}=\frac{\operatorname{LCS}(Q, Y)}{\operatorname{len}(Y)}
\end{equation}
\begin{equation}
    P_{l c s}=\frac{\operatorname{LCS}(Q, Y)}{\operatorname{len}(Q)}
\end{equation}
\begin{equation}
    F_{l c s}=\frac{\left(\beta^{2}+1\right) R_{l c s} P_{l c s}}{R_{l c s}+\beta^{2} P_{l c s}}
\end{equation}
where $R_{l c s}$ represents the recall, $P_{l c s}$ represents the accuracy, $\operatorname{LCS}(Q, Y)$ represents the common subsequence of the model generation question and the gold question, and $\beta$ is the fixed parameter.

\subsubsection{Question Generation}
This paper compares the S2S+ATT+CP model and TiBERT, S2S+ATT+CP is a seq2seq model with the attention and copy mechanism  model \cite{b29}. We use TiBERT as the embedding to the traditional seq2seq architecture, and use LSTM as the decoding, the parameters of the TiBERT-based question generation model are shown in Tab \ref{tab5}.

From Tab \ref{tab6}, we can see that all indicators of TiBERT are higher than the baseline model, the BLEU-2 of TiBERT reaches 28.60$\%$, which is 8.46$\%$ higher than the BLEU-2 of S2S+ATT+CP. On the value of ROUGE-L, TiBERT is 8.59$\%$ higher than S2S+ATT+CP. This shows that the performance of TiBERT model is better than S2S+ATT+CP in question generation task. The main reason is that TiBERT can generate contextual vector representations, which is useful for the most tasks of Tibetan downstream architecture.

Finally, we conduct a case analysis of the generated questions, we compare the questions generated by the two models with the gold questions, as shown in Tab \ref{tab7}. We highlight the differences between the generated questions and the gold questions. For the first question,two models generate the same question as the gold question. For the second question, the TiBERT model generates the same question as the gold question, but there is a duplication of "elevation" in the question generated by the S2S+ATT+CP model, and the "meters" were missing. For the third question, the TiBERT model also generates the same question as the gold question, but there was a duplication of "southwest" in the question generated by the S2S+ATT+CP model, and the "kilometers" in the gold question is changed to "centimeters". For the fourth question, there is a duplication of "many" in the question generated by the TiBERT model, but the question generated by the S2S+ATT+CP model is completely wrong. The main reason is that the answer to the question is a number, and the model is difficult to learn useful knowledge. For the fifth question, the words "Red Star" is missing and “commune” is repeated three times in the question generated by the S2S+ATT+CP model, the question generated by the TiBERT model is a declarative sentence, which is the answer of the gold question, which shows that the understanding ability of TiBERT is not enough and there is still much room for improvement in question generation task of TiBERT.

We can conclude that TiBERT model performs better than the S2S+ATT+CP model on the generation task, which shows that the TiBERT model can be applied to the generation task. In addition, since we only use the word embedding information, there will be some word errors in the generated question, in the subsequent research, we can use TiBERT to fine-tune to generate better questions.

\section{Conclusion} 
This paper trains a pre-trained language model named TiBERT for Tibetan, and we verify the effect of TiBERT on the two downstream tasks of text classification and question generation. On the text classification task, we verify the performance of TiBERT on long text and short text. Compared with CNN and the minority language pre-training model, TiBERT obtains the best classification effect. On the question generation task, we compare TiBERT with the sequence-to-sequence model. The experimental results show that TiBERT achieves better results, which shows that our model is effective. At the same time, it verifies that the monolingual pre-training model is better than the multilingual pre-training language model.

\section*{Acknowledgment}

This work is supported by National Nature Science Foundation (No. 61972436).

\end{document}